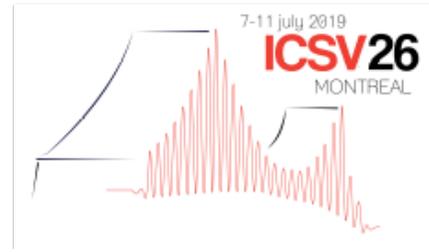

# QUESTIONNAIRE ANALYSIS TO DEFINE THE MOST SUITABLE SURVEY FOR PORT-NOISE INVESTIGATION

Andrea Cerniglia
*Accon Italia Srl, San Genesio ed Uniti (Pv), Italia*
*e-mail: andrea.cerniglia@accon.it*

Davide Chiarella, Paola Cutugno, Lucia Marconi
*CNR-ILC, Genova, Italia*
*e-mail: davide.chiarella@ilc.cnr.it, paola.cutugno@ilc.cnr.it, lucia.marconi@ilc.cnr.it*

Anna Magrini, Gelsomina Di Feo
*DICAr,Università degli Studi di Pavia, Pavia, Italia*
*email: magrini@unipv.it, gf.difeo@gmail.com*

Melissa Ferretti
*CNR-IEIIT, Genova, Italia*
*e-mail: melissa.ferretti@ieiit.cnr.it*

The high level of noise pollution affecting the areas between ports and logistic platforms represents a problem that can be faced from different points of view. Acoustic monitoring, mapping, short-term measurements, port and road traffic flows analyses can give useful indications on the strategies to be proposed for a better management of the problem. A survey campaign through the preparation of questionnaires to be submitted to the population exposed to noise in the back-port areas will help to better understand the subjective point of view. The paper analyses a sample of questions suitable for the specific research, chosen as part of the wide database of questionnaires internationally proposed for subjective investigations. The preliminary results of a first data collection campaign are considered to verify the adequacy of the number, the type of questions, and the type of sample noise used for the survey. The questionnaire will be optimized to be distributed in the TRIPLO project (TRansports and Innovative sustainable connections between Ports and LOgistic platforms). The results of this survey will be the starting point for the linguistic investigation carried out in combination with the acoustic monitoring, to improve understanding the connections between personal feeling and technical aspects.

Keywords: port noise, acoustic monitoring, subjective survey, psychoacoustics

## 1. Introduction

In environmental assessments, purely technical research can objectively describe a phenomenon, but does not guarantee its universality with respect to its perception. What then moves to complete a choice or to express a judgment is the result of a process that involves different levels of subjectivity, not necessarily linked to the development of a linear neurological type, but of a neuropsychological one, and therefore not concretely measurable.

The same acoustic phenomenon can be perceived differently by the same individual or by several individuals, in relation to his state or their states. It follows that a sound can be considered both as a physical phenomenon, and therefore measurable through objective parameters, and as a phenomenon linked to sound perception, of a subjective nature and consequent to the subject's psycho-physical-emotional state. These two characteristics are strictly interdependent, so it is often insufficient to simply examine them separately.

It should also be noted that the parameters aimed at estimating noise levels are generally obtained with technical methods and, through the use of adequate measuring instruments. These procedures refer to regulatory limits, defined by empirical studies and aimed at protecting a wide range of potentially exposed citizens. Very often the reference thresholds do not correspond to the actual individual perception: therefore the technical investigation alone may not lead to a complete knowledge of the problem, not representing the real conditions of perception.

The population is a valuable source of information in assessing the quality of the space in which they live or work, suggesting the relationships between environment, comfort and productivity. For these reasons it becomes useful to use subjective investigation tools, through which the opinion of the population can become a valid support for traditional methods of analysis, and can improve the overall evaluation.

The multiple investigation techniques are distinguished on the basis of the methodology of administration: direct (e.g. face-to-face) or indirect interviews (telephone, on websites, etc.), questionnaires, direct observations (carried out only by the surveyor), diaries (special types of questionnaires specifically designed to record frequent or unimportant events such as low expenses or daily activities), mixed techniques (e.g. diary and interview).

To get surveys, for large-scale research, on situations related to living habits, the best survey formula is the questionnaire. In fact it allows to collect direct information between the researcher and the respondent, so that they are easily comparable. To obtain reliable data it is good that the survey is formulated following a procedural scheme, as defined for example in [1]; it follows that the processing phases are strictly dependent on the planning of the sampling processes and the preparation of the applications.

The project developed is aimed at assessing the acoustic impact on the population exposed to noise generated by retro-port activities, in relation to individual perception. To achieve the objectives mentioned above, the work involves various phases and involves different sample sites.

Specifically, it is considered to proceed with the combined analysis of acoustic data from sound detection sensors installed in the various areas and results obtained through the administration of questionnaires to the population exposed to the events.

This will make it possible to create integrated interactive maps, obtained by associating the words derived from the elaboration of the questionnaires to the sound acquisitions. The analyses carried out on heterogeneous samples will allow to identify and, possibly, report the most relevant correlations in terms of subjective perception, so that the descriptive and characteristic parameters of noise in the external environment are more understandable to the population.

For the realization and development of the activities related to the questionnaires, acoustic records and measurements were made in the areas of: Genova, Lucca, Livorno, Sassari, Porto Torres and Toulon; these records were made in places with different sound characteristics in which the simultaneous evaluation of the conventional acoustic parameters was carried out.

The acquired data will be processed both in terms of extraction of significant samples of specific sounds, and in terms of calculation of spectra and other descriptors, including psychoacoustic ones.

## 2. State of the art: acoustic questionnaires and psychoacoustics aspects

The phenomenology related to acoustic perception is particularly complex: it involves, in addition to basic parameters of intensity, and frequency, also other characteristics such as: duration, repetitiveness, temporal variability.

In order to describe the noise from a perceptive point of view, psychoacoustics introduces a series of new specific descriptors such as loudness (strength), roughness (roughness), sharpness (sharpness), fluctuation strength, and others [2]. These descriptors therefore also take into consideration the temporal development of the various components that constitute the considered noise.

Although this discipline lends itself to objectively describing perceptive phenomena, it does not provide information on the actual induced sensation, which otherwise can be obtained by investigating the opinion of the population regarding specific sound events. To this end, subjective investigations can be structured in different ways, depending on the type of phenomenon to be assessed. In fact, they not only play a fundamental role in the correlation between individual judgments and data obtained through measurements, but also provide information related to collateral aspects with respect to the research area only.

The acoustic environment depends on the sources present in the territory: transport infrastructures, industries, anthropic sources, and natural noises. Usually, it is complex to determine, according to the perceptive aspect, the contribution of the different sources. For this reason, several researches (for example [3], [4]), relating to noise in the external environment, have relied on the use of subjective investigation tools to obtain more representative results of the examined situation.

## 3. Subjective survey for retro-port areas: development and methodology

The noise associated with port and retro-port infrastructures is usually characterized by specific sources, such as ships and everything related to them (engines, chimneys, sirens, etc.), and also involves other contexts related to industry, commerce, tourism that are instead characterized by other sources of noise (forklifts and cranes for loading and unloading goods, road and / or rail traffic, container handling, etc.).

To evaluate the disturbance generated by this type of noise, typical of some urban contexts, it was decided to carry out a subjective investigation, to support the usual acoustic measurements, through the distribution of a questionnaire.

In order to develop a tool that can be easily compiled by the population exposed to noise in the areas examined in the project, a methodology structured in two distinct phases has been developed. Specifically, a first "prototype" questionnaire was created and distributed (in Italian language), to a "pilot" sample and characterized by a structure of questions mainly consisting of open-ended questions. The second phase, currently being developed, will provide for a further "final" questionnaire to be distributed in the territories involved in the survey. In this article the methods of elaboration and administration of only the prototype questionnaire will be described.

The preliminary investigation is aimed at creating a database of frequently used terms related to acoustic phenomena characteristic of the external environment and, specifically, of retro-port contexts. It is made up of a part of questions regarding personal data (i.e. age, gender etc.) followed by sections in which cognitive processes are pursued through the phases of listening to sounds / noises and are strictly aimed at knowing the opinion of the interviewees with respect to the field of investigation. As already mentioned, part of the audio included in the questionnaire derives from recordings made in the different areas and therefore related to contexts and activities that take place in the areas studied. Other sounds/noises proposed are instead identifiers of sound phenomena that can occur in everyday life, to support the realization of a wider vocabulary.

The priority areas identified for the realization of the prototype questionnaire are those connected to the implementation of the structure and to the connections between audio signals and words.

With reference to the first area, the intent has been to develop a smooth and rapidly implementable structure obtained through the use of the "Google Forms" service of the Google platform. The prototype questionnaire consisted of a very essential record section that was used in part to structure and organize information in a simple way in relation to the different questions asked in the questionnaires.

The audio files included in the questionnaire via the YouTube platform, were grouped into macro areas according to the following criteria: external environment, daily environment and natural environment. The choice of adjectives to be proposed as options for answering some closed questions was conducted using information obtained from linguistic analyses carried out on a reference corpus.

## 4. Structure of the prototype questionnaire

In the following paragraph the structure of the questionnaire is described and details are provided with respect to the corpus and linguistic analyses taken as a reference, as well as some information relating to the selected adjectives.

Following the recording activities carried out in the various areas, the sound tracks have been divided into the following assessment areas shown below.

1. Daily environment: leaky faucet, open faucet, bells, clock.

2. Natural environment: crickets, gulls, sea, rain, wind in the leaves.

3. Port and retro-port environment: rail crossing, container handling, boat ropes handling, plastic squeaking, ship trumpet, forklift, crane, chainsaw.

4. Ship docking in the port of Porto Torres, Livorno and Toulon; vehicular traffic in Sassari, Lucca and Toulon; truck at Porto Torres, Lucca and Toulon, Train near Lucca, Genoa and generic location (specific scope of the project).

The draft questionnaire was articulated according to the following criteria: use of simple language; formulation of clear and targeted questions, carried out on a single scope; absence of ambiguity in the terms and in the questions structure (for example the use of the word "audio" is preferred to "noise" or "sound").

As already mentioned, the objective of the questionnaire is a collection of sample terms, referable to all the sound tracks identified. In order to keep the attention of the respondent high, it was decided to use a short and easily understandable compilation structure. According to the foregoing reasons, given the large number of audio samples available, it was decided to distribute them in four separate groups (i.e. A, B, C, D), avoiding incurring a very long and laborious questionnaire. The questionnaire was structured with 5 sections with questions, according to the following schema:

1. Info Section: personal information
2. "ThreeSources (TS)" Section: after listening three audios belonging to the same source category, the respondent had to declare if the proposed soundtracks were to be considered similar or different. The four types of sources are respectively "Docking Ship", "Traffic", "Camion" and "Train". Based on the response acquired, the activation of the relative sub-section was envisaged:
   - Section "ThreeSources_similar (TS_s)": in case of reply "Similar" in the previous section, it was requested to identify the source with a noun and to describe it with one or more adjectives;
   - Section "ThreeSources_different (TS_d)": in case of reply "Different" it was requested to identify the audio tracks listened to, by three nouns and, subsequently, to describe it with an adjective.
3. Question Section (Q): it was proposed to listen to an audio related to a single sound source which was asked to match adjectives, nouns, verbs and colours.

4. Source Section (S): one of the audio presented in the previous TS section was played. The question in this case was based on closed answers; more specifically it was asked to make a choice from one or more sets of selected terms, presented through one or more "clouds" of words:

- Source 1 word cloud (S1): a single cloud of terms (example in Figure 1) containing selected adjectives, according to the procedures described above, was proposed; the respondent was asked to choose at least three representative adjectives to describe the recording.
- Source 4 word clouds (S4): similar to question of type S1, but four clouds of adjectives were proposed. The four clouds match the categories previously defined, according to criteria adopted in other researches [5,6,7]:
  *a. Sensation / Evaluation: adjectives related to the perceptive evaluation of the source;*
  *b. Environment / Space: adjectives that describe the spatial characteristics;*
  *c. Time / Time variability: adjectives that describe the temporal characteristics;*
  *d. Intensity / Strength: adjectives that describe the intensity of the source*

In Figure 2 the organization of the five sections is shown, while in Table 1 the sources chosen for the four questionnaire groups (A,B,C,D) are indicated.

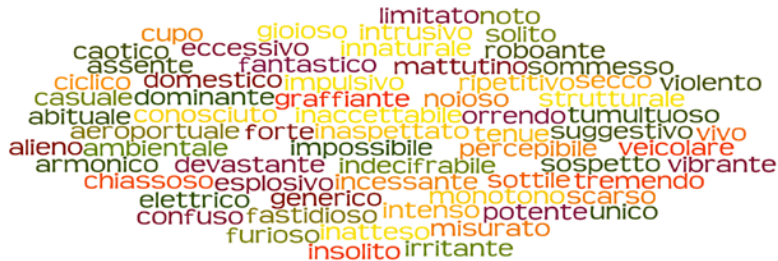

Figure 1: Example of the proposed clouds of terms

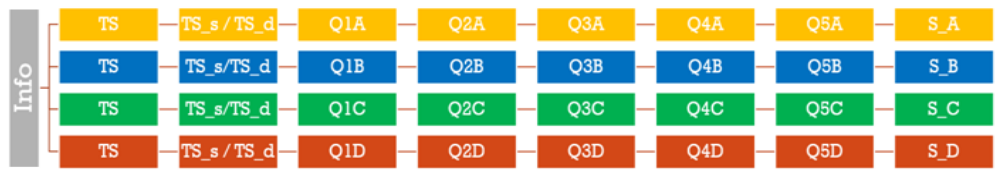

Figure 2: Organization of the sections in relation to the four groups of questionnaires

Table 1: indications of the sources chosen in relation to the four proposed questionnaires

| Sect. | Questionnaire A | Questionnaire B | Questionnaire C | Questionnaire D |
|---|---|---|---|---|
| TS | Attracco (docking) | Traffico veicolare (traffic) | Camion (truck) | Treno (train) |
| Q1 | Passaggio a livello (rail crossing) | Sirena nave (ship's siren) | Muletto(forklift) | Gru (tower crane) |
| Q2 | Grilli (crickets) | Gabbiani (seagulls) | Cigolio plastico (plastic squeak) | Motosega (chainsaw) |
| Q3 | Container (containers) | Corde (ropes) | Rubinetto aperto (open faucet) | Pioggia (rain) |
| Q4 | Mare calmo (calm sea) | Pioggia (rain) | Gabbiani (seagulls) | Campane (bells) |
| Q5 | Rubinetto che perde (defective faucet) | Rubinetto che perde (defective faucet) | Vento tra le foglie (wind) | Orologio (clock) |
| S | Attracco Tolone (docking) | Traffico Tolone (traffic) | Camion P. Torres (truck) | Treno Genova (train) |

## 4.1 Information from corpora and construction of adjective clouds for Source Section

The linguistic corpora consist of collections, usually of large dimensions, of oral or written texts. They are fundamental both for being able to recognize and classify a given linguistic phenomenon, both to carry out studies on the evolution of the language and / or contribute to the construction of linguistic analysis tools, by deriving the information from empirical data.

For the construction of the adjective cloud inserted in Source Section S adjectives were extracted from the Italian Web 2016® corpus (itTenTen16), composed of written texts collected on the Web.

The texts that constituted the corpus were and subjected to a set of linguistic analyses of lexical and grammatical type through 'Sketch Engine' [8, 9], a platform able to construct, explore and analysing corpora. This system also allows the generation of "word sketches", that is the combinations of lexical words and / or co-occurrences based on the analysed corpus (i.e. Italian Web 2016) [10, 11]. From the available results it was therefore proceeded to the extraction of the words connected to the word "noise", and on them subsequent elaborations were carried out.

In particular, for the analyses, 442 adjectives associated with the noun "noise" have been extracted from the corpus, ordered by occurrences and divided according to the characteristics shown in the previous descriptive sections (i.e. S1, S4).

Given the large number of adjectives extracted, although a subsequent revision was carried out and those considered most pertinent were selected, all adjectives could hardly be represented in a single graph. Therefore, in each branch of the questionnaire, only a subset of adjectives was presented, adopting as a criterion the one based on the number of occurrences of appearance in the corpus: the most frequent has been attributed to branch1, the second most frequent to branch2, and so on up to branch4 and subsequently, the attribution of the adjectives is started again following the branch1, etc.

## 5. Data collection and processing method

Once the structure and organization of the questionnaire were defined, it was implemented through the service provided by the 'Google Forms' platform.

The platform used for the survey distribution, whereas on the one hand simplified the creation of the questionnaire and the preliminary analysis of the data, on the other hand was not exhaustive for the subsequent elaborations envisaged by the survey. For this reason, a pre-processing of the results was required, characterized by a "purification" phase of the answers from possible anomalies or inconsistencies due to an incorrect interpretation of the requests or typing errors. This allowed the creation of a first database, obtained using the SQLLite library, which contains the various occurrences of each adjective, noun, verb, and colour.

### 5.1 Analysis of the first results of the prototype questionnaire

The questionnaire was administered to a pilot sample that was found to be like 402 respondents (55.6% male; 43.4% female; 1% undeclared), not necessarily inhabitants in the areas under investigation. The pilot sample, also called "control group", is usually used as a reference point, as it is considered neutral with respect to the variables of interest. It was then decided to administer the prototype questionnaire to a group of participants who were not directly involved in the retro-port contexts in order to have a further term of comparison, and to understand the differences between actual and perceived noises / sounds.

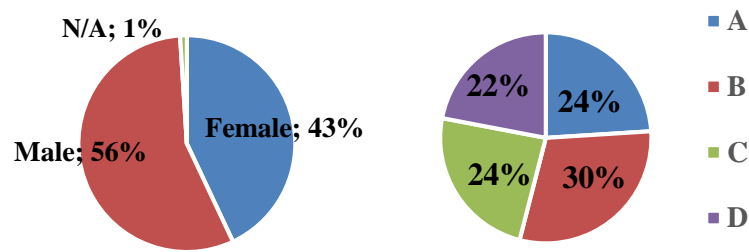

Figure 3: Distribution by gender of respondents and by occurrences of questionnaires of type A, B, C, D

Figure 3 shows respectively the distribution of the genus related to the sample investigated and the distribution of the respondents to the 4 branches of the questionnaire.

In this phase, the data processing concerned only the evaluation of the adjectives associated with the sound tracks related to open-ended and closed-answer questions. The four sections 'TS' in Table 2 show the answers and their percentages of occurrence provided by the interviewees.

Table 2: Percentage results referring to adjectives chosen by the interviewees

| ATTRACCO (docking) | | TRAFFICO (traffic) | | CAMION (truck) | | TRENO (train) | |
|---|---|---|---|---|---|---|---|
| Adjective | % | Adjective | % | Adjective | % | Adjective | % |
| Rumoroso (noisy) | 29 | Fastidioso (annoying) | 39 | Rumoroso (noisy) | 29 | Fastidioso (annoying) | 39 |
| Ripetitivo (repetitive) | 16 | Rumoroso (noisy) | 30 | Ripetitivo (repetitive) | 16 | Rumoroso (noisy) | 30 |
| Forte (loud) | 11 | Forte (loud) | 11 | Forte (loud) | 11 | Forte (loud) | 11 |
| Metallico (metallic) | 6 | Rilassante (relaxing) | 6 | Metallico (metallic) | 6 | Rilassante (relaxing) | 6 |
| Cupo (dark) | 5 | Ripetitivo (repetitive) | 6 | Cupo (dark) | 5 | Ripetitivo (repetitive) | 6 |
| Rimbombante (resounding) | 5 | Urbano (urban) | 6 | Rimbombante (resounding) | 5 | Urbano (urban) | 6 |

The results show a tendency towards the definition of "annoying" and "noisy" for all the audios included in the questionnaire. It should also be noted that the generic adjective "strong" is chosen in all the case studies proposed, appearing, in some cases, in third place in terms of number of occurrences. It is in a consistent line with the information extracted from the reference linguistic corpus, where it is identified as the second most used adjective and associated with the word "noise". However, the "Relaxing" outlier remains to be noted in the cases of "Traffic" and "Trucks".

The proposal of a cloud of terms, selected according to the criteria already described and inserted in the closed answer sections, provided results reported in Table 3.

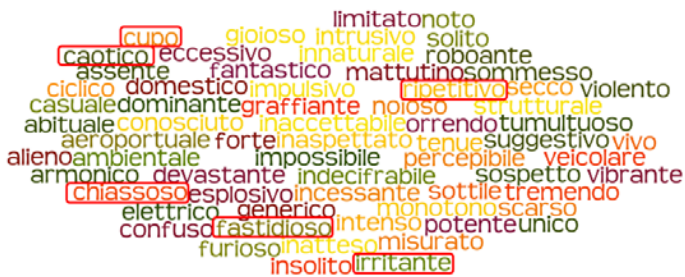

Figure 4: Cloud of adjectives proposed for section S1_A (docking) with the first six adjectives highlighted with greater number of occurrence

Table 3: Occurrences for closed-ended questions

| Adjective | Occurrence (%) |
|---|---|
| Ripetitivo (repetitive) | 28% |
| Irritante (irritating) | 25% |
| Fastidioso (annoying) | 21% |
| Caotico (chaotic) | 20% |
| Cupo (dark) | 20% |
| Chiassoso (rowdy) | 18% |

Although the proposed adjectives vary in the most disparate meanings, as reported in the example in Figure 4, audios have been mainly associated with terms that identify disturbing and annoying conditions. Differently, it is noticeable from the previous case of free associations, that the term "strong", although present among the proposed adjectives, is not a favourite word among respondents.

# 6. Conclusions

From an initial examination it emerges that the proposed audio tracks and, specifically, those relating to the project area, have been associated with terms that identify a negative meaning of the sensation transmitted during listening. Therefore, being a subjective investigation carried out on a 'control group', not directly affected by the characteristic sound phenomena of the port and retro-port areas, it emerges that the condition of annoyance or disturbance generated by the investigated sources is univocally perceived, regardless of the specific involvement of the interviewee.

The words' extraction, conducted taking into account both linguistic and acoustic aspects, will constitute the database which will then contain the response options to be included in the final questionnaire, administered to the population exposed in the port areas and retro-port, with closed questions.

Data processing is still under development: therefore the present considerations are to be considered preliminary and not exhaustive. Deeper evaluations regarding the relationships between acoustic parameters and selected terms (i.e. adjectives) are planned and, in a second step, similar analyses on colours, verbs and nouns will be carried out.

## AKNOWLEDGMENTS

The research is carried out within the framework of the TRIPLO project financed by the In-terreg Cooperation Program V-A Italy France Maritime 2014 - 2020. Priority axis 3 - Improvement of the connection of the territories and of the sustainability of port activities. Specific objective 7C1 - Improve the sustainability of commercial ports and logistic platforms associated to them, in the scope to the reduction of noise pollution. The use of the Sketch Engine platform was made possible thanks to the European project H2020 Elexis (grant agreement No 731015).